\documentclass{article}

\PassOptionsToPackage{numbers}{natbib}
\usepackage[preprint]{neurips_2025}

\usepackage{hyperref}
\usepackage{url}
\usepackage{array}
\usepackage{multirow}
\usepackage{amsmath}
\usepackage{amssymb}
\usepackage{mathtools}
\usepackage{amsthm}
\usepackage{booktabs}

\title{Learning Neuron Dynamics within Deep Spiking Neural Networks}

\theoremstyle{plain}

\theoremstyle{definition}

\theoremstyle{remark}

\newcommand{\R}{\mathbb{R}}

\author{
  Eric~Jahns \\
  STAM Center\\
  Arizona State University\\
  Tempe, Arizona \\
  \texttt{jjahns@asu.edu} \\
  \And
  Davi Moreno \\
  Center for Advanced Studies and \\
  Systems of Recife\\
  \texttt{dcma@cesar.org.br} \\
  \And
  Michel A. Kinsy \\
  STAM Center\\
  Arizona State University\\
  Tempe, Arizona \\
  \texttt{mkinsy@asu.edu} \\
}

\begin{document}

\maketitle

\begin{abstract}
Spiking Neural Networks (SNNs) offer a promising energy-efficient alternative to Artificial Neural Networks (ANNs) by utilizing sparse and asynchronous processing through discrete spike-based computation. However, the performance of deep SNNs remains limited by their reliance on simple neuron models, such as the Leaky Integrate-and-Fire (LIF) model, which cannot capture rich temporal dynamics. While more expressive neuron models exist, they require careful manual tuning of hyperparameters and are difficult to scale effectively. This difficulty is evident in the lack of successful implementations of complex neuron models in high-performance deep SNNs.

In this work, we address this limitation by introducing Learnable Neuron Models (LNMs). LNMs are a general, parametric formulation for non-linear integrate-and-fire dynamics that learn neuron dynamics during training. By learning neuron dynamics directly from data, LNMs enhance the performance of deep SNNs. We instantiate LNMs using low-degree polynomial parameterizations, enabling efficient and stable training. We demonstrate state-of-the-art performance in a variety of datasets, including CIFAR-10, CIFAR-100, ImageNet, and CIFAR-10 DVS. LNMs offer a promising path toward more scalable and high-performing spiking architectures.
\end{abstract}

\section{Introduction}

Artificial Neural Networks (ANNs) have seen widespread adoption due to their remarkable success in areas such as computer vision \citep{JunyiChai-2021} and natural language processing \citep{WahabKhan-2023}, among others. However, this success has come with a substantial increase in energy consumption~\citep{KashuYamazaki-2022}. In response, Spiking Neural Networks (SNNs) have emerged as a promising energy-efficient alternative. Unlike ANNs, which rely on continuous, synchronous computations, SNNs operate asynchronously using discrete, sparse events known as spikes. This spiking behavior is typically driven by biologically inspired neuron models~\citep{WulframGerstner-2014}. The event-driven and spiking nature of SNNs enables them to mimic the brain’s sparse connectivity and energy-efficient behavior.

When mapped onto hardware designed to exploit these characteristics, SNNs can achieve significantly lower energy consumption than traditional ANNs~\citep{NitinRathi-2023}. Such hardware, known as neuromorphic hardware, is designed to support the sparse and event-driven computation style of SNNs. Notable examples include IBM’s TrueNorth chip \citep{FilippAkopyan-2015}, used for always-on speech recognition in edge devices \citep{Wei-YuTasi-2017}, and Intel’s Loihi 2 \citep{IntelLoihi2}, which enables ultra-low-power image classification tasks~\citep{GregorLenz-2023}.

In deep spiking neural networks, the temporal evolution of a neuron's membrane potential is typically governed by a fixed neuron model. The Leaky Integrate-and-Fire (LIF) neuron model, defined by a linear leakage of membrane potential over time, has become the standard due to its simplicity and stability~\citep{WulframGerstner-2014, YujieWu-2018}. However, the expressiveness of the LIF model is limited, as it cannot capture the diverse range of nonlinear temporal behaviors observed in biological neurons. More sophisticated neuron models—such as the Quadratic Integrate-and-Fire, Exponential Integrate-and-Fire, and Adaptive Exponential models—offer richer dynamics and enhanced modeling capacity. Despite their theoretical advantages, these models introduce added complexity and additional hyperparameters (e.g., thresholds, adaptation currents, non-linear scaling factors) that are difficult to tune and often require domain-specific knowledge. As a result, their use in deep SNN architectures remains limited, with only the LIF model demonstrating consistent success in scalable, high-performance models.

The reliance on simple neuron models, such as LIF, can limit a network's performance, as they fail to capture the rich temporal structure required for complex tasks. Motivated by this limitation, we advocate for a shift from simple neuron dynamics and propose Learnable Neuron Models (LNM), in which a neuron's dynamics are parameterized and optimized alongside the rest of the network. By making a neuron's dynamics task-adaptive, learnable neuron models enhance the performance of SNNs without the difficulties associated with integrating sophisticated neuron models. To summarize, the main contributions of our work are as follows:

\begin{itemize}
    \item We derive a general parametric formulation that encompasses the family of nonlinear integrate-and-fire neuron models, allowing arbitrary differentiable functions to be integrated into the membrane dynamics and learned during training. We call our formulation \textbf{Learnable Neuron Models (LNM)}.
    \item We realize \textbf{Learnable Neuron Models} through low-degree polynomial parameterizations, enabling neuron dynamics to be learned directly from data. By leveraging Horner's method for polynomial evaluation, our approach introduces negligible energy overhead compared to LIF.
    \item Our method achieves state-of-the-art accuracy compared to LIF-based convolutional deep SNNs on both static and neuromorphic datasets, with results of $97.01\%$ on CIFAR-10, $80.70\%$ on CIFAR-100, and $81.39$\% on CIFAR10-DVS using ResNet-19 and $70.91$\% on ImageNet using ResNet-34 surpassing the previous best results by $0.54\%$, $0.50\%$, $1.55$\%, and $0.17$\%, respectively.
\end{itemize}

\section{Related Work}

\subsection{Deep Learning with Spiking Neural Networks}

Training spiking neural networks directly with backpropagation is challenging. This is mainly due to the non-differentiability of spiking behavior. In recent years, several techniques have been developed to overcome this issue. One of the most popular techniques is called surrogate gradients. Instead of calculating the exact gradient of the spiking behavior found in SNN, surrogate gradients provide an approximation that can be used to perform backpropagation. Training performance largely depends on the quality of this approximation. Additionally, to make training SNNs with surrogate gradients stable and improve their performance, various techniques have been developed. These include developing new loss functions~\citep{ShikuangDeng-2022, YufeiGuo-2022, HangchiShen-2024}, normalization techniques~\citep{HanleZheng-2020, ChaotengDuan-2022, YufeiGuo-2023}, and learning optimal surrogate gradients~\citep{YuhangLi-2021, ShuangLian-2023, ShikuangDeng-2023}.

We want to note that, due to the lack of support for training on neuromorphic datasets when using ANN-to-SNN conversion techniques~\citep{YongqiangCao-2015}, we restrict all comparisons in this work to direct training techniques with surrogate gradients.

\subsection{Neuron Models and Parameter Learning}

Several works that utilize direct training techniques, such as \citet{WeiFang-2021, XingtingYao-2022, NitinRathi-2023, ShuangLian-2023, ShuangLian-2024}, look to modify the LIF neuron to increase model performance. For example, \citet{WeiFang-2021} proposes a trainable decay factor of the LIF neurons that is independently learned for each layer. \citet{NitinRathi-2023} builds upon this idea by utilizing a learnable threshold in addition to the decay factor. \citet{XingtingYao-2022} proposed modifications to the LIF neuron, which make it act similar to long-short term memory by utilizing a gating mechanism that can choose the optimal biological features during training. \citet{ShuangLian-2023} proposes dynamically adjusting the surrogate gradient window during training when using a learnable decay factor to minimize gradient mismatch. \citet{ShuangLian-2024} proposes a temporal-wise attention mechanism, allowing neurons to establish connections with past temporal data selectively. Finally, \citet{YufeiGuo-2024} aims to reduce the information loss of binary spiking behavior by utilizing ternary spikes, i.e., spikes with negative magnitude. Additionally, the authors parameterize their ternary spike with a learnable constant, allowing neurons to fire real-valued spikes.

In this work, we propose enhancing the performance of SNNs by learning the dynamics of neurons in each layer of a network. To the best of our knowledge, this area of SNNs has not yet been explored.

\section{Background}
\subsection{Spiking Neural Networks}

While ANNs use continuous-valued data to transmit information, SNNs use discrete events, known as spikes. The choice of neuron model governs the spiking dynamics of an SNN. In modern deep SNN research, non-linear integrate-and-fire neuron models are typically preferred for their blend of computational efficiency and biologically plausible dynamics \citep{WulframGerstner-2014}. The general form for all non-linear integrate-and-fire neuron models is
\begin{align}
\label{eq:nonlinear_integrate_and_fire_ode}
\tau \frac{du}{dt} = f(u) + RI.
\end{align}
where $\tau$ is a membrane time constant, $u$ is the membrane potential, the function $f(u)$ governs the intrinsic dynamics of $u$, $R$ is a linear resistor, and $I$ is pre-synaptic input. By changing the function $f(u)$, we obtain different neuron models within the non-linear integrate-and-fire family. For example, if we define $f(u) = u_{r} - u$, for some resting potential $u_{r}$, we obtain the LIF neuron model.

To utilize non-linear integrate-and-fire neuron models in deep learning scenarios, discretization is required~\citep{ChaotengDuan-2022}. The most commonly used discretization technique is Euler's method~\citep{YujieWu-2018}. Applying Euler's method to Equation~\eqref{eq:nonlinear_integrate_and_fire_ode}, we obtain
\begin{align}
\label{eq:nonlinear_integrate_and_fire_discrete}
u(t+1) = u(t) + \frac{\Delta t}{\tau}(f(u(t)) + RI(t)),
\end{align}
where $t$ is the timestep and $\Delta t$ is the discretization step size. To simplify Equation~\eqref{eq:nonlinear_integrate_and_fire_discrete}, we can fold the term $\frac{\Delta t}{\tau} R$ into the pre-synaptic weights of $I$ to obtain
\begin{align}
\label{eq:nonlinear_integrate_and_fire_discrete_simplified}
u(t+1) = u(t) + \frac{\Delta t}{\tau}f(u(t)) + I(t).
\end{align}

When $u(t+1)$ exceeds a threshold $u_{th}$, a spike is produced and the voltage is reset as follows,
\begin{align}
    \label{eq:NLIAFUpdate1}
    u(t+1) &= [u(t) + \frac{\Delta t}{\tau}f(u(t))](1 - o(t)) + I(t) \\
    \label{eq:NLIAFUpdate2}
    o(t+1) &= \Theta(u(t+1) - u_{th}),
\end{align}
where $\Theta$ is the Heaviside function given by $\Theta(x) = 0$ if $x < 0$, else $\Theta(x) = 1$. Equations~\eqref{eq:NLIAFUpdate1} and~\eqref{eq:NLIAFUpdate2} enable the efficient implementation of forward and backward propagation in spatial and temporal domains~\citep{HanleZheng-2020}.

\subsubsection{Training Spiking Neural Networks}

The Spatial-Temporal Back Propagation (STBP) algorithm is commonly used to train SNNs~\citep{YujieWu-2018}. First, an SNN processes temporal data for $T$ timesteps. The SNN's output is decoded by accumulating the synaptic voltage of the last layer as follows
\begin{align}
    \hat{y} = \frac{1}{T} \sum_{t=1}^T W o(t).
\end{align}
In the above equation, $W$ is a weight matrix, $o(t)$ is the spiking activity at timestep $t$, and $\hat{y} \in \mathbb{R}^m$ is the model output given $m$ classes. Using our output vector $\hat{y} = (\hat{y}_1, \hat{y}_2, \ldots, \hat{y}_m)$ and a label vector $y = (y_1, y_2, \ldots, y_m)$, we compute the cross entropy loss, $L$, between $\hat{y}$ and $y$. Then, using the STBP algorithm, we train the network using the chain rule to update synaptic weights by
\begin{align}
\label{eq:stbp}
\frac{\partial L}{\partial W_n} &= 
\sum_{t=1}^T 
\Bigg[
    \frac{\partial L}{\partial o_n(t\!+\!1)} 
    \frac{\partial o_n(t\!+\!1)}{\partial u_n(t\!+\!1)} +
    \frac{\partial L}{\partial u_n(t\!+\!1)}
    \frac{\partial u_n(t\!+\!1)}{\partial u_n(t)}
    \Bigg] \frac{\partial u_n(t+1)}{\partial W_n},
\end{align}

where $n$ is the layer of the network, $u(t)$ is a neuron's membrane potential, and $o(t)$ is a neuron's spike output~\citep{YufeiGuo-2023}.

To overcome the undefined derivative, $\frac{\partial o_n(t+1)}{\partial u_n(t+1)}$, \citet{YujieWu-2018} proposed using the derivative of an approximation to the Heaviside function with useful gradient information. This technique is known as a surrogate gradient. A common choice of surrogate gradient is the rectangle function \citep{HanleZheng-2020, ShikuangDeng-2022, ShuangLian-2023}, which is defined by
\begin{align}
    \label{eq:RectangleSG}
    \frac{\partial o(t+1)}{\partial u(t+1)} \approx \frac{1}{\alpha} \mathrm{sign}\left (|u(t+1) - u_{th}| < \frac{\alpha}{2}\right ),
\end{align}
where $\alpha$ determines the width and area of the surrogate gradient and typically remains constant throughout training. The choice of $\alpha$ has a significant impact on the learning process of SNNs, with improper choices leading to gradient mismatches and approximation errors.

\section{Methodology}

\subsection{Limitations of Fixed Neuron Models}

In deep spiking neural networks, a single function $f(u)$ is typically chosen to govern the membrane dynamics of all neurons in the network. The LIF neuron model is the most widely adopted due to its computational efficiency and simplicity. Although more expressive models—such as the quadratic integrate-and-fire, exponential integrate-and-fire, and adaptive exponential integrate-and-fire neurons—are available, they require careful manual tuning of hyperparameters, which often vary across tasks and datasets. In practice, identifying stable and effective hyperparameters for training deep SNNs with such neuron models remains a significant and unsolved challenge. This problem is evident as, to date, only the LIF model has shown success in high-performance deep SNNs. 

\subsection{Learnable Neuron Models}

To address this issue, we propose a parameterization of a neuron's dynamics with a set of learnable weights $\theta \in \R^N$, yielding a function $f_\theta(u)$. Specifically, each layer of an SNN can learn a unique set of weights $\theta$, thereby increasing the network's expressiveness and enabling it to learn the optimal neuron models for a given task.

Incorporating $f_\theta(u)$ into Equation~\eqref{eq:nonlinear_integrate_and_fire_discrete} provides us with
\begin{align}
\label{eq:nonlinear_integrate_and_fire_discrete_lnm}
u(t+1) = u(t) + \frac{\Delta t}{\tau}f_\theta(u(t)) + I(t).
\end{align}
We can then fold the constant factor $\frac{\Delta t}{\tau}$ into $\theta$, resulting in a simplified update rule,
\begin{align}
\label{eq:nonlinear_integrate_and_fire_discrete_lnm_simplified}
u(t+1) = u(t) + f_\theta(u(t)) + I(t).
\end{align}

\subsubsection{Choice of Parameterization}

A key choice in our methodology lies in selecting a suitable parameterization for $f_\theta$, which defines the subthreshold dynamics of the neuron. This choice is guided by three primary constraints: computational efficiency, expressiveness, and differentiability. We evaluate four function families under these criteria: multi-layer perceptrons (MLPs)~\citep{Perceptron}, splines~\citep{Splines}, Chebyshev polynomials~\citep{Chebyshev}, and polynomials.

\textit{MLPs} are arguably the most expressive choice of function. Their universal approximation capability enables them to represent highly nonlinear and task-specific dynamics with a relatively small number of parameters. However, MLPs are not suitable for neuromorphic hardware due to their continuous-valued, synchronous data processing.

\textit{Splines} provide a compelling middle ground between MLPs and polynomials. Cubic splines, in particular, can offer piecewise smooth approximations with local control, which could prove advantageous when modeling neuron dynamics that exhibit distinct regimes of behavior. However, in our testing, splines introduced the most significant computational overhead due to the difficulty in efficiently vectorizing their operations.

\textit{Chebyshev Polynomials} offer a promising trade-off among the three constraints; however, low-degree variants, necessary for computational efficiency, tend to exhibit undesirable sinusoidal behavior due to their orthogonal basis. This, in turn, constrains the dynamics of a neuron. Additionally, Chebyshev polynomials require their input to be scaled. This can be a costly operation and not one typically suited to neuromorphic processors~\citep{IntelLoihi2}.

\textit{Polynomials} contrast with Chebyshev polynomials in that low-degree polynomials don't necessarily exhibit a specific behavior. While polynomials may require high degrees to obtain the same expressivity as the other function families, they are the most computationally efficient. Additionally, polynomials align with the dynamics of existing neuron models, such as the LIF and QIF. Consequently, we adopt polynomials as the most practical parameterization, defining $f_\theta$ as
\begin{align}
\label{eq:polynomial_parameterization}
f_\theta(u(t)) &= \sum_{i=0}^N \theta_i u(t)^i.
\end{align}

Our use of polynomial functions introduces two key challenges. First, while polynomials can offer arbitrary expressiveness dependent on their degree, low-degree polynomials are more favorable for computational efficiency. This, in turn, limits the expressiveness of $f_\theta$. Second, polynomial functions exhibit unbounded growth outside a limited input range, which can potentially lead to instability during training or inference. To mitigate both of these issues, we utilize low-degree polynomials and clip the input of $f_\theta$ to the range of $[-1, 1]$. Clipping ensures that values below $-1$ and above $1$ are set to $-1$ and $1$, respectively. This bounding, in turn, improves the numerical stability of polynomials. Working within this interval also increases the expressiveness of polynomials, as we focus on a small region to learn the dynamics of neurons, allowing better utilization of low-degree polynomials.

\subsubsection{Training}

To train the learnable weights of LNM using STBP, we use our updated neuron logic in Equation~\eqref{eq:nonlinear_integrate_and_fire_discrete_lnm_simplified} and define the derivative of a weight $\theta_k$ with respect to a loss function $L$ as
\begin{equation}
\label{eq:lnm_grad}
\begin{aligned}
\frac{\partial L}{\partial \theta_k} = \sum_{t=1}^T \Bigg[
&\frac{\partial L}{\partial o_n(t\!+\!1)} 
  \frac{\partial o_n(t\!+\!1)}{\partial u_n(t\!+\!1)} + \frac{\partial L}{\partial u_n(t\!+\!1)} 
  \frac{\partial u_n(t\!+\!1)}{\partial f_\theta(u_n(t))}\Bigg]
\frac{\partial f_\theta(u_n(t))}{\partial \theta_k}
\end{aligned}
\end{equation}

In the equation above, $n$ is the network layer, and $u(t)$ and $o(t)$ are the neuron's membrane potential and spike output at timestep $t$, respectively. Additionally, we need to redefine the gradient of synaptic weight $W_n$ as

\begin{equation}
\label{eq:new_weight_grad}
\begin{aligned}
\frac{\partial L}{\partial W_n} &= \sum_{t=1}^T 
\Bigg[
    \frac{\partial L}{\partial o_n(t\!+\!1)} 
    \frac{\partial o_n(t\!+\!1)}{\partial u_n(t\!+\!1)} +
    \frac{\partial L}{\partial u_n(t\!+\!1)} 
    \left(1 + \frac{\partial f_\theta(u_n(t))}{\partial u_n(t)}\right)
\Bigg] 
\frac{\partial u_n(t+1)}{\partial W_n}.
\end{aligned}
\end{equation}

Using Equations~\eqref{eq:lnm_grad} and~\eqref{eq:new_weight_grad}, STBP can be implemented with our proposed learnable neuron model as long as $f_\theta$ is differentiable.

\subsubsection{Efficient Polynomial Evaluation}

To efficiently evaluate the polynomial $f_\theta$, we utilize Horner's method~\citep{HornersMethod}. That is, we iteratively factor $u(t)$ terms out of Equation~\eqref{eq:polynomial_parameterization}. For example, we can evaluate a third-degree polynomial using Horner's method as follows:
\begin{align}
    \label{eq:naive_poly_eval}
    f_\theta(x) &= \theta_0 + \theta_1 u(t) + \theta_{2} u(t)^2 + \theta_{3} u(t)^3 \\
    \label{eq:horner_poly_eval}
    &= \theta_0 + u(t)(\theta_1 + u(t)(\theta_2  + \theta_3 u(t)))
\end{align}

Horner's method enables evaluation of an $N$-degree polynomial in $\mathcal{O}(N)$ time, compared to the $\mathcal{O}(N^2)$ time required by na\"ive evaluation.

If $f_\theta$ is at least a first-degree polynomial, then we can further simplify Equation~\eqref{eq:nonlinear_integrate_and_fire_discrete_lnm_simplified}. That is, for an N-degree polynomial, with $N \geq 1$, observe
\begin{align}
    u(t+1) &= u(t) + f_\theta(u(t)) + I(t) \notag \\
           &= u(t) + \theta_0 + \theta_1 u(t) + \cdots + \theta_{N} u(t)^N + I(t) \notag \\
           &= \theta_0 + (1 + \theta_1) u(t) + \cdots + \theta_{N} u(t)^N + I(t) \notag \\
           &= f_\theta(u(t)) + I(t).
\end{align}
This simplification removes one addition otherwise required to evaluate $u(t+1)$.

Post-training, techniques such as least-squares polynomial approximation~\citep{LeastSquaresPolynomialApproximations} can be used to find a lower-degree polynomial to approximate the learned polynomial and further reduce computational complexity. This can be applied in scenarios when an LNM has been overparameterized or when weight regularization techniques are applied.

\section{Experiments}

We test our proposed learnable neuron model and compare it to recent state-of-the-art works. We utilize widely-adopted model architectures and datasets, such as ResNet-19~\citep{HanleZheng-2020} for CIFAR-10 and CIFAR-100 \citep{CIFAR10_CIFAR100}, ResNet-34~\citep{HanleZheng-2020} for ImageNet, and ResNet-19~\citep{HanleZheng-2020} and VGGSNN~\citep{ShikuangDeng-2022} for CIFAR-10 DVS~\citep{CIFAR10_CIFAR100}. For all experiments, we compare the average top-$1$ validation accuracy for each model across three training runs. To promote stability during training, we initialize each LNM as the LIF neuron model and constrain $f_\theta(0) = 0$ for all choices of $\theta$. Lastly, we adopt a similar hyperparameter choice as done in the recent work of \citet{HangchiShen-2024}. Our exact hyperparameters and dataset augmentations are detailed in Appendix~\ref{appendix:hyperparam}.

\subsection{Ablation Study}

\begin{figure}[h]
  \centering
  \includegraphics[width=0.75\linewidth]{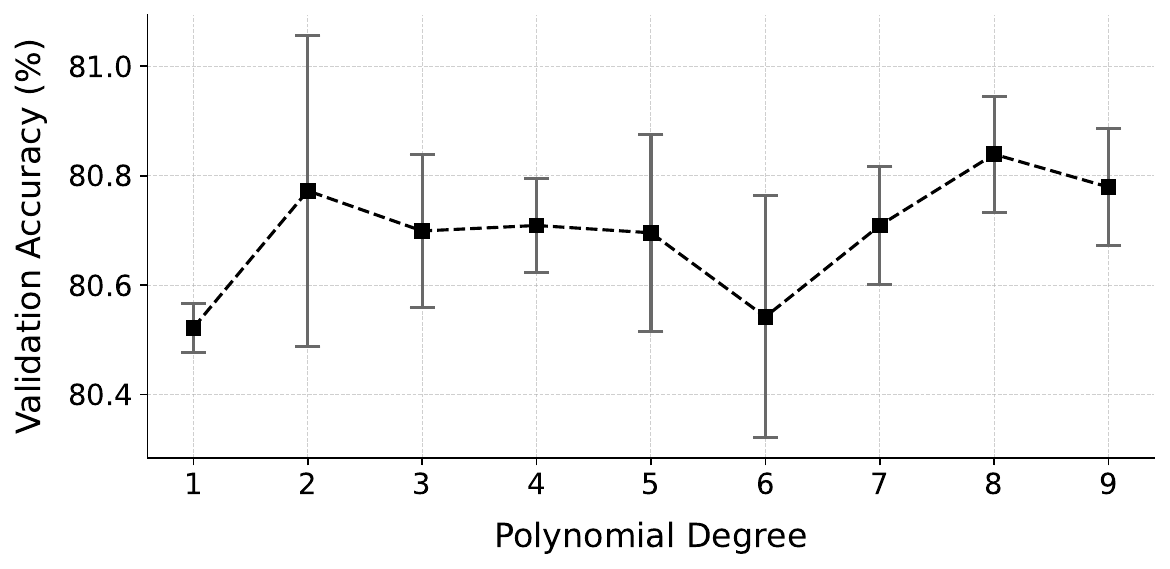}
  \caption{Ablation study on polynomial degree.}
    \label{fig:ablation_study}
\end{figure}

To determine the optimal polynomial degree, we conduct an ablation study using a ResNet-19 architecture trained on CIFAR-100 with four timesteps. For each polynomial degree, we perform three independent training runs and report the mean and standard deviation to assess stability and performance. The results, summarized in Figure~\ref{fig:ablation_study}, show that our learnable neuron model can benefit from higher-degree polynomials. However, we also observe increased variability as the degree increases. Based on these results, we select third-degree polynomials as the optimal choice, offering a favorable balance between computational cost, model stability, and accuracy.

\begin{table*}[t!]
\caption{Summary and comparison of accuracy results on static datasets. The \textbf{Timesteps} column correlates to \textbf{CIFAR-10}/\textbf{CIFAR-100}/\textbf{ImageNet}. All \textbf{CIFAR-10}/\textbf{CIFAR-100} results utilize ResNet-19 while all \textbf{ImageNet} results utilize ResNet-34. Acronyms: Surrogate Gradient (SG), Learnable Neuron Model (LNM). Initial Membrane Potential (IMP).}

\vspace{1em}

\centering
\scriptsize

\resizebox{\linewidth}{!}{
\begin{tabular}{lrcccc}
    \toprule
    \textbf{Work} & \textbf{Method} & \centering \textbf{Timesteps} & \centering \textbf{CIFAR-10} & \centering \textbf{CIFAR-100} & \centering\arraybackslash \textbf{ImageNet} \\
    \midrule
     TEBN \cite{ChaotengDuan-2022} & Batch Normalization & \centering 4/4/4 & \centering 94.71\% & \centering 76.41\% & \centering\arraybackslash 68.28\% \\
     TET \cite{ShikuangDeng-2022} & Loss Function & \centering 6/6/6 & \centering 94.50\% & \centering 74.72\% & \centering\arraybackslash 68.00\%\\
     IM-Loss \cite{YufeiGuo-2022} & Loss Function + SG & \centering 2/2/6 & \centering 93.85\% & \centering 70.18\% & \centering\arraybackslash 67.43\% \\
     LocalZO + TET \cite{BhaskarMukoty-2023} & Direct Training & \centering 2/2/$\times$ & \centering 95.03\% & \centering 76.36\% & \centering\arraybackslash $\times$ \\
     Surrogate Module \cite{ShikuangDeng-2023} & Hybrid & \centering 4/4/4 & \centering 95.54\% & \centering 79.18\% & \centering\arraybackslash 68.25\% \\
     MPBN \cite{YufeiGuo-2023} & Membrane Normalization & \centering 2/2/4 & \centering 96.47\% & \centering 79.51\% & \centering\arraybackslash 64.71\% \\
     Backpropagation Shortcuts \cite{YufeiGuo-2024-Shortcut} & Direct Training & \centering 2/2/4 & \centering 95.19\% & \centering 77.56\% & \centering\arraybackslash 67.90\% \\
     IMP + LTS \cite{HangchiShen-2024} & IMP + Loss Function & \centering $\times$/$\times$/4 & \centering $\times$ & \centering $\times$ & \centering\arraybackslash 68.90\% \\
     \cmidrule{1-6}
     \multirow{1}{*}{PLIF \cite{WeiFang-2021}} & \multirow{1}{*}{Neuron Model} 
           & \centering 8/$\times$/$\times$ & \centering 93.50\% & \centering\arraybackslash $\times$ & \centering\arraybackslash $\times$ \\
    \cmidrule{2-6}
    \multirow{2}{*}{GLIF \cite{XingtingYao-2022}} & \multirow{2}{*}{Neuron Model} 
           & \centering 4/4/4 & \centering 94.85\% & \centering\arraybackslash 77.05\% & \centering\arraybackslash 67.52\% \\
     &  & \centering 2/2/$\times$ & \centering 94.44\% & \centering\arraybackslash 75.48\%  & \centering\arraybackslash $\times$ \\ 
     \cmidrule{2-6}
     \multirow{2}{*}{LSG \cite{ShuangLian-2023}} & \multirow{2}{*}{Neuron Model + SG}  
           & \centering 4/4/$\times$ & \centering 95.17\% & \centering\arraybackslash 76.85\% & \centering\arraybackslash $\times$ \\ 
     &  & \centering 2/2/$\times$ & \centering 94.41\% & \centering\arraybackslash 76.32\% & \centering\arraybackslash $\times$ \\ 
     \cmidrule{2-6}
     IM-LIF \cite{ShuangLian-2024} & Neuron Model + Loss Function & \centering 3/3/$\times$ & \centering 95.29\% & \centering 77.21\% & \centering\arraybackslash $\times$ \\
     \cmidrule{2-6}
     Ternary Spike \cite{YufeiGuo-2024} & Neuron Model & \centering 2/2/4 & \centering 95.80\% & \centering 80.20\% & \centering\arraybackslash 70.74\% \\
     \cmidrule{2-6}
     \multirow{2}{*}{\textbf{LNM (Ours)}} & \multirow{2}{*}{\textbf{Neuron Model}} 
           & \centering \textbf{4}/\textbf{4}/\textbf{4} & \centering \textbf{97.01 $\pm$ 0.04\%} & \centering \textbf{80.70 $\pm$ 0.14\%} & \centering\arraybackslash \textbf{70.91\% $\pm$ 0.03\%}\\ 
     &  & \centering \textbf{2}/\textbf{2}/$\times$ & \centering \textbf{96.96 $\pm$ 0.10\%} & \centering \textbf{80.07 $\pm$ 0.13\%} & \centering\arraybackslash $\times$ \\
    \bottomrule
\end{tabular}
}
\label{table:results}
\end{table*}

\begin{table*}[t!]
\caption{Comparison between state-of-the-art techniques and our Learnable Neuron Model (LNM) on CIFAR-10 DVS.}
\vspace{1em}
\centering
\scriptsize
\resizebox{\linewidth}{!}{
\begin{tabular}{llrrcc}
    \toprule
    \textbf{Dataset} & \textbf{Work} & \textbf{Method} & \textbf{Architecture} & \centering \textbf{Timesteps} & \textbf{Accuracy} \\
    \midrule
    \multirow{10}{*}{CIFAR-10 DVS}
     & TET \cite{ShikuangDeng-2022} & Loss Function &  VGGSNN & \centering 10 & \centering\arraybackslash 77.40\% \\
     & IM-Loss \cite{YufeiGuo-2022} & Loss Function + SG & ResNet-19 & \centering 10 & \centering\arraybackslash 72.60\% \\
     & LocalZO + TET \cite{BhaskarMukoty-2023} & Direct Training & VGGSNN & \centering 10 & \centering\arraybackslash 75.62\% \\
     & MPBN \cite{YufeiGuo-2023} & Membrane Normalization & ResNet-19 & \centering 10 & \centering\arraybackslash 74.40\% \\
    \cmidrule{2-6}
     & GLIF \cite{XingtingYao-2022} & Neuron Model & ResNet-19 & \centering 16 & \centering\arraybackslash 78.10\% \\
     & LSG \cite{ShuangLian-2023} & Neuron Model + SG & VGGSNN & \centering 10 & \centering\arraybackslash 77.90\% \\
     & IM-LIF \cite{ShuangLian-2024} & Neuron Model + Loss Function & VGGSNN & \centering 10 & \centering\arraybackslash 80.50\% \\
     & Ternary Spike \cite{YufeiGuo-2024} & Neuron Model & ResNet-19 & \centering 10 & \centering\arraybackslash 79.84\% \\
     \cmidrule{3-6}
     & \multirow{2}{*}{\textbf{LNM (Ours)}} & \multirow{2}{*}{\textbf{Neuron Model}} & \textbf{VGGSNN} & \centering \textbf{10} & \centering\arraybackslash \textbf{82.95 $\pm$ 0.69\%} \\
     & & & \textbf{ResNet-19} & \centering \textbf{10} & \centering\arraybackslash \textbf{81.39 $\pm$ 0.25\%} \\
    \bottomrule
\end{tabular}
}
\label{table:results_cont}
\end{table*}

\subsection{Comparison to Recent Works}

First, we compare our method to the state-of-the-art (SOTA) on CIFAR-10. Looking at Table~\ref{table:results}, the best previous result achieves $96.47$\% accuracy. Our LNM approach surpasses this, reaching $96.96$\% and $97.01$\% with two and four timesteps, respectively. On CIFAR-100, our method achieves $80.07$\% with two timesteps and $80.70$\% with four timesteps. The prior best result reports $80.20\%$ with two timesteps~\citep{YufeiGuo-2024}. 

We next evaluate our LNM approach on the ImageNet dataset. Our technique achieves an accuracy of $70.91\%$, surpassing the previous state-of-the-art Ternary Spike method of \citet{YufeiGuo-2024} by $0.17\%$. Compared to the best-performing binary spike method by \citet{HangchiShen-2024}, our approach achieves a $2.01\%$ higher accuracy. These results demonstrate the scalability of our method.

Finally, we evaluate our method on CIFAR-10 DVS using VGGSNN and ResNet-19. In Table \ref{table:results_cont}, we see VGGSNN trained with our method achieves $82.95$\% accuracy.  Our method outperforms the previous best result of $80.50$\% by $2.45$\%, achieving a notable gain. Using ResNet-19, we improve on the prior state-of-the-art by $1.55$\%, achieving $81.39$\%. These results affirm that incorporating more expressive neuron dynamics can significantly enhance the performance of deep SNNs. Note that we do not compare against the work of \citet{ChaotengDuan-2022}. Their method introduces Temporal Effective Batch Normalization, which cannot be folded into model weights after training. This can result in significant overhead as it removes the ability to perform event-driven inference on neuromorphic hardware.

\subsection{Learned Neuron Model Analysis}

\begin{figure*}[t!]
  \centering
  \includegraphics[width=\linewidth]{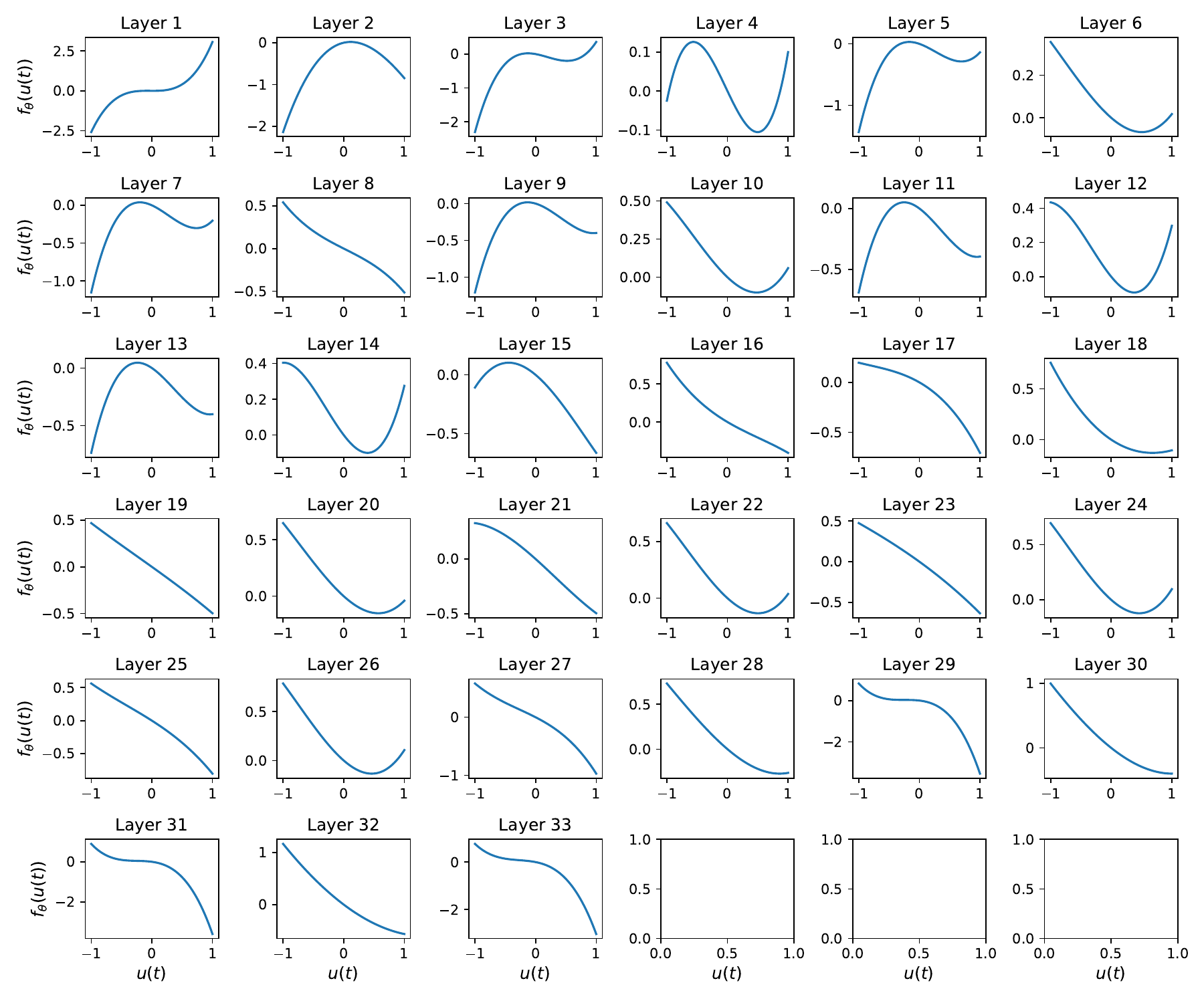}
  \caption{Learned neuron models of ResNet-34 trained on ImageNet with four timesteps.}
    \label{fig:cifar100_neuron_models}
\end{figure*}

We examine the learned neuron models within a ResNet-34 architecture, as shown in Figure~\ref{fig:cifar100_neuron_models}. This figure illustrates the distinct neuron models learned across each spiking layer of the network. Noting that we utilized a threshold of $0.5$ for these experiments, only a small portion of layers (8, 10, 19, 21, 23, 25) learned neuron models that resemble LIF dynamics. Layers 6, 12, 14, 20, 22, 24, and 26 all learned dynamics similar to Quadratic Integrate-and-Fire (QIF) neurons~\citep{WulframGerstner-2014}. A few layers (2, 7, 9, 11, 13, and 15) learned dynamics that resemble the behavior of negative QIF neurons, a phenomenon not observed in biological neurons~\citep{WulframGerstner-2014}. Other layers (18, 28, 30, 32) interestingly learned negative Exponential Integrate-and-Fire (ExLIF) behavior~\citep{WulframGerstner-2014}. The remaining layers all learned dynamics that don't resemble any known neuron models. For example, layer 4 learned sinusoidal behavior, while layers 1, 3, 5, 29, 31, and 33 learned cubic behavior. Layers 16, 17, and 27 demonstrate logarithmic behavior. This diversity suggests that deep SNNs do benefit from a richer variety of neuron behaviors.

\subsection{Energy Efficiency}

We calculate the average hardware energy cost per inference of LNMs. Note that the first layer of the network employs floating-point operations (FLOPS), while the remaining layers utilize synaptic operations (SOPS). We follow the technique used by \citet{YufeiGuo-2024} and calculate the energy cost of each layer $E_i$ as
\begin{align}
    \label{eq:SNNEnergyConsumption}
    E_i = T \cdot (fr \cdot E_\text{AC} \cdot OP_\text{AC} + E_\text{MAC} \cdot OP_\text{MAC}).
\end{align}
In the above equation, $T$ is the number of timesteps, $fr$ is the average spike-rate of neurons in layer $i$, $OP_\text{AC}$ and $OP_\text{MAC}$ are the number of accumulations and multiply-and-accumulate operations, respectively, with $E_\text{AC}$ and $E_\text{MAC}$ being their corresponding energy cost. We assume these operations take place with $32$-bit floating point values on $45$nm technology where $E_\text{MAC} = 4.6pJ$ and $E_\text{AC} = 0.9pJ$~\citep{ShuangLian-2024}. We compare the energy cost of both LNM and the LIF neuron in Table~\ref{table:energy_cost}, assuming the same spike rate for both. It can be seen that LNM only introduces energy consumption overheads between $2 - 5.5$\% compared to LIF. 

\begin{table}[h]
    \caption{Approximate energy consumption for third-degree LNM and LIF models. ResNet-19 was used for CIFAR-10, CIFAR-100, and CIFAR-10 DVS. ResNet-34 was used for ImageNet.}
    \vspace{0.75em}
    \centering
    \begin{tabular}{l c r r r}
        \toprule
        \textbf{Dataset} & \textbf{Timesteps} & \textbf{LIF Energy (mJ)} & \textbf{LNM Energy (mJ)} & \textbf{LNM Overhead} \\
        \midrule
        CIFAR-10      & 2  & 0.547 & 0.570 & 4.20\%\\
        CIFAR-100     & 2  & 0.649 & 0.672 & 3.54\%\\
        CIFAR-10 DVS  & 10 & 4.758 & 5.015 & 5.40\%\\
        ImageNet      & 4  & 13.725 & 14.020 & 2.15\%\\
        \bottomrule
    \end{tabular}
    \label{table:energy_cost}
\end{table}

\section{Limitations}

One limitation of LNMs is the added computational complexity they bring compared to the LIF neuron model. In our experiments, this added complexity increases the energy consumption of SNNs trained with LNMs between $2-5.5$\% compared to the LIF neuron. While the energy consumption overhead can be lowered by decreasing the polynomial degree, it requires extra consideration of its potential trade-off in performance.

Our choice of low-degree polynomials may limit the expressiveness and variability of learned neuron models. Additionally, clipping the input to polynomials to the range $[-1, 1]$ can cause information loss, as values outside this range become indistinguishable. Therefore, finding alternatives to polynomials that don't significantly increase energy consumption, computational cost, or decrease model stability is important within future research.

Another limitation of our work is that it only considers single-variable non-linear integrate-and-fire neuron models. Multi-variable neuron models can present more complex and biologically plausible dynamics, albeit at the cost of increased computational complexity. Exploring how these multi-variable neuron models can be incorporated into our LNM methodology to improve performance further remains an interesting future research direction.

Lastly, our formulation of LNMs introduces challenges in identifying optimal surrogate gradient parameters. While \citet{ShuangLian-2023} proposes a method to analytically derive optimal surrogate gradient windows for LIF neurons based on the statistical properties of the neuron, extending this approach to LNMs is non-trivial.

\section{Conclusion}

In this work, we introduced Learnable Neuron Models (LNMs) as a scalable and efficient solution to the limitations of traditional neuron models in deep SNNs. By learning expressive, non-linear dynamics directly from data, LNMs enable more adaptable and high-performing spiking architectures without requiring manual tuning or sacrificing training stability. LNMs demonstrate state-of-the-art performance for convolutional SNNs on several benchmark datasets, including CIFAR-10, CIFAR-100, ImageNet, and CIFAR-10 DVS, showcasing their effectiveness as an alternative to traditionally used neuron models.

\bibliographystyle{plainnat}
\bibliography{refs}

\newpage

\appendix

\section{Hyperparameters and Experimental Setup}
\label{appendix:hyperparam}

Table~\ref{tab:hyperparams} details the hyperparameters used for each dataset and experiment.

\begin{table*}[h!]
    \centering
    \begin{tabular}{lrrrr}
    \toprule
         Dataset & CIFAR-10 & CIFAR-100 & CIFAR-10 DVS & ImageNet \\
         \midrule
         Architecture & ResNet-19 & ResNet-19 & VGGSNN/ResNet-19 & ResNet-34 \\
         Timesteps & 2/4 & 2/4 & 10 & 4 \\
         Surrogate Gradient & Rectangle  & Rectangle & Triangle & Rectangle \\
         Epochs & 400 & 400 & 200 & 350 \\
         Optimizer & SGD & SGD & SGD & SGD \\
         Scheduler & Cosine & Cosine & Cosine & Cosine \\
         Learning Rate & 1e-1 & 1e-1 & 1e-1 & 1e-0 \\
         LNM Learning Rate & 1e-1 & 1e-1 & 1e-2 & 1e-2 \\
         Momentum & 9e-1 & 9e-1 & 9e-1 & 9e-1 \\
         Weight Decay & 5e-4 & 5e-4 & 5e-4 & 5e-5\\
         Label Smoothing & 1e-1 & 1e-1 & 1e-2 & 0e-0\\
         Warm Up Epochs & 0 & 0 & 0 & 5 \\
         A100 80GB GPUs & 1 & 1 & 1 & 4 \\
         \bottomrule
    \end{tabular}
    \caption{Hyperparameters for each dataset and model architecture.}
    \label{tab:hyperparams}
\end{table*}

\subsection{Dataset Augmentations}

We detail the data augmentations applied to each dataset used within our work. Note that for all datasets besides CIFAR-10 DVS, we normalize by the mean and standard deviation. For CIFAR-10~\citep{CIFAR10_CIFAR100}, we utilize cropping, flipping, and Cutout~\citep{Cutout}. For CIFAR-100~\citep{CIFAR10_CIFAR100}, we similarly utilize cropping, flipping, and Cutout~\citep{Cutout} with the addition of AutoAugment~\citep{AutoAugment}. For CIFAR-10 DVS~\cite{CIFAR10DVS}, we resize each frame to $48\times48$ and apply random cropping and flipping. Finally, for ImageNet~\cite{ImageNet}, we utilize random cropping and flipping.

\end{document}